\documentclass{article}

\usepackage[preprint, nonatbib]{neurips_2024}

\usepackage{wrapfig}
\usepackage{graphicx} 
\usepackage{amsmath}

\usepackage[
backend=biber,
sorting=ynt
]{biblatex}
\usepackage[utf8]{inputenc} 
\usepackage[T1]{fontenc}    
\usepackage{hyperref}       
\usepackage{url}            
\usepackage{booktabs}       
\usepackage{amsfonts}       
\usepackage{nicefrac}       
\usepackage{microtype}      
\usepackage{xcolor}         

\title{Diffusion-Based Generation of Neural Activity \newline from Disentangled Latent Codes}

\author{
Jonathan D. McCart$^{1,2}$\thanks{Contact: jmccart6@gatech.edu, chethan@gatech.edu
}  \quad Andrew R. Sedler$^{1,2}$\thanks{Current address: AE Studio} \quad Christopher Versteeg$^2$ \\ \textbf{Domenick Mifsud}$^{1,2}$ \quad \textbf{Mattia Rigotti-Thompson}$^{2}$ \quad
\textbf{Chethan Pandarinath}$^{1,2,3*}$ \\
$^1$Center for Machine Learning, Georgia Tech \\ $^2$Department of Biomedical Engineering, Georgia Tech and Emory University \\
$^3$Department of Neurosurgery, Emory University School of Medicine 
}

\begin{document}

\maketitle

\begin{abstract}
 Recent advances in recording technology have allowed neuroscientists to monitor activity from thousands of neurons simultaneously. Latent variable models are increasingly valuable for distilling these recordings into compact and interpretable representations. Here we propose a new approach to neural data analysis that leverages advances in conditional generative modeling to enable the unsupervised inference of disentangled behavioral variables from recorded neural activity. Our approach builds on InfoDiffusion, which augments diffusion models with a set of latent variables that capture important factors of variation in the data. We apply our model, called Generating Neural Observations Conditioned on Codes with High Information (GNOCCHI), to time series neural data and test its application to synthetic and biological recordings of neural activity during reaching. In comparison to a VAE-based sequential autoencoder, GNOCCHI learns higher-quality latent spaces that are more clearly structured and more disentangled with respect to key behavioral variables. These properties enable accurate generation of novel samples (unseen behavioral conditions) through simple linear traversal of the latent spaces produced by GNOCCHI. Our work demonstrates the potential of unsupervised, information-based models for the discovery of interpretable latent spaces from neural data, enabling researchers to generate high-quality samples from unseen conditions.
\end{abstract}

\section{Introduction}
\vspace{-0.2cm}
New neural interfaces enable the simultaneous recording of hundreds to thousands of neurons \cite{recording_affects_analysis, Neuropixels}. Interpreting the resulting high-dimensional datasets is a major challenge, as neural activity may have non-linear and time-varying relationships to many variables of interest. Thus neuroscientists have begun to rely on latent variable models to distill high-D recordings of neural population activity into compact and more easily interpretable representations. Of particular interest are generative latent variable models, which allow neuroscientists to map back from the latent variables to the high-D population activity itself to study how variables of interest are geometrically encoded within neural activity.

Brains can produce a rich behavioral repertoire, making it challenging to know what variables are encoded by a particular brain region or neural sub-population. Traditionally, neuroscientists assess how variables are encoded in a brain region by designing experiments that constrain behavior to a small set of predetermined variables of interest. An alternative approach uses latent variable models to discover which variables are encoded in a neural circuit and how those variables are encoded at the level of individual neurons. While auxiliary variables (observed behavior) may be used to structure latent spaces \cite{piVAE,schneider2023cebra}, such approaches are not possible when key variables are unobserved or unknown. To promote more general discovery of latent structure, models would ideally have the following properties: unsupervised application to complex and diverse time-series data, structured latent spaces that are disentangled with respect to behavioral variables of interest, and queryable, conditional generation of high-quality samples for unseen behavioral conditions. 

We propose a new approach to neural data analysis that builds on advances in conditional generative models to enable the unsupervised inference of disentangled behavioral variables from recorded neural activity. Our approach uses InfoDiffusion \cite{InfoDiffusion}, which augments diffusion models with a set of latent variables (here termed “codes”) that capture important factors of variation in the data. By maximizing the mutual information between the observed and latent variables, InfoDiffusion has been shown to learn disentangled representations of human-interpretable latent factors.

We extend InfoDiffusion to sequential data and test its performance on synthetic and biological datasets from neural circuits performing a motor task. We compare our approach, Generating Neural Observations Conditioned on Codes with High Information (GNOCCHI), to an alternative unsupervised generative latent variable model for neural time-series data, latent factor analysis via dynamical systems (LFADS). While both models achieve high-quality generation of neural data, GNOCCHI achieves higher-quality latent spaces that are more clearly structured with respect to behavioral variables. Moreover, the latent spaces produced by GNOCCHI are more disentangled than those produced by LFADS, such that simple linear traversal of the latent space along identified paths produces novel samples with isolated changes in their relationship to behavior. Finally, we show that these properties enable high-quality conditional generation of neural activity for unseen behavioral conditions.

\section{Related Work}
\vspace{-0.2cm}

Generative models have frequently been used to uncover latent, low-dimensional descriptions of neural population activity, including models based on Gaussian processes \cite{GPFA,GP_factor_models,zhao2017variational,wu2017gaussian}, variational autoencoders (VAEs) \cite{gao2016linear,LFADS_original} including those using self-supervised learning \cite{SwapVAE}, neural ordinary differential equations \cite{PLNDE,sedler2022expressive,ODIN}, and recent diffusion-based approaches\cite{vetter2023generating}. In this work we chose a sequential autoencoder, latent factor analysis via dynamical systems (LFADS \cite{LFADS_original}), as our point of comparison, as it reduces time series neural data to a latent representation whose structure is determined by a standard VAE objective \cite{AEVB}. LFADS allows samples to be generated for novel conditions through a simple interpolation approach we detail in Section \ref{sec: latent-navigation}. We note an alternate point of comparison could be SwapVAE \cite{SwapVAE}, which decomposes its latent space into qualitatively different latents for “style” and “content”; in this case we chose LFADS due to its frequent applications to neural time series data \cite{LFADS_NatMethods,AutoLFADS,zhu2022deep}.

Prior work in interpretable deep generative models explicitly conditions generation on behavioral labels \cite{piVAE, JointNeuroBehavior}. These approaches may allow identifiable and interpretable recovery of latent structure for observed variables, but are not applicable when key variables are unobserved or unknown.

Generative models have been used in other domains to produce samples from outside of the training set without access to the underlying variables of interest. This has been achieved with success on image datasets through modifying the generative cost of VAEs \cite{beta-VAE}. Another popular class of approaches has been to regularize the mutual information between the observed data and the latent space of the generator \cite{InfoVAE, InfoDiffusion}. Through interpolation and extrapolation in the latent space, these models can generate samples that reflect an interpretable transition between variables of interest. However, to our knowledge these information-based deep generative models -- which do not rely on behavioral labels -- have not previously been applied to neuroscientific datasets.

\section{Methods \label{sec: methods}}
\vspace{-0.2cm}
\subsection{Artificial neural datasets \label{sec: artificial-neural-activity}}
\vspace{-0.2cm}

To generate realistic neural activity on which to test our models, we trained an RNN (128-hidden unit GRU), to perform planar reaching tasks using a 2-link, 6 muscle arm model \cite{MotorNet, CtD_Benchmark}. The RNN model received a 17-dimensional input: endpoint coordinates (2D), muscle lengths (6D) and velocities (6D), target coordinates (2D) and a go cue input (1D). The loss function of this model was to minimize the mean-squared error between the effector endpoint and the target, while minimizing the weighted squared muscle activation. This training paradigm produced a biologically plausible set of neural activations with associated task parameters that can be used as a benchmark for evaluating the latents/codes produced by the different methods. More details on the inputs and outputs of the RNN are shown in Figure \ref{fig:SyntheticFull}A, with full training details given in Section \ref{appendix: task-trained} of the Appendix. 

We consider a random target task, where the RNN must control the set of 6 muscles to manipulate the effector endpoint from a start location to a target location, both of which are randomly selected from a $6 \times 6$ grid. At a random time during the trial, a target was presented, followed by a delayed go cue that instructed the RNN to initiate movement towards the presented target. 

To create a training dataset for our generative models, we generated 1000 total trials of RNN activity from each task. We then aligned to a window around the go cue time, producing trials of length 0.4 seconds (40 bins). When assessing the generalization of the representations learned by the generative models, we would remove trials that had specific target locations. For the random target task, we held out trials corresponding to target locations on the middle square of grid points (see Figure \ref{fig:SyntheticHoldout}A). These trials were used as a "held-out" dataset to quantify how well our models could generate data from conditions that were not in their training set. Additional details can be found in Tables \ref{tab:full_data-trial-counts} and \ref{tab:heldin-heldout-trial-counts}.

\subsection{Biological neural datasets \label{sec: biological-neural-activity}}
\vspace{-0.2cm}
To assess whether our approach works for biological neural recordings, we used previously published recordings from a multi-electrode array \cite{UtahArray} implanted in monkey primary motor cortex (M1) during a self-paced random target reaching task \cite{ODoherty_RTT_2017}. Voltage signals from each of the 96 electrodes were converted into threshold crossings, binned at 20 ms, and aligned to a window [-200, 500] ms around movement onset. When assessing the generalization capability of the generative models, we held-out reaches to targets in the upper-right quadrant ($\sim$14\% of the trials, see Fig. \ref{fig:NeuralData}). We analyze the generation quality of these data when testing the ability of our models to generate accurate examples of neural activity from outside its training set. For modeling this dataset, we used spiking activity with LFADS, and smoothed spiking activity with GNOCCHI (see Section \ref{lfads-smoothing} for more details).

\subsection{Model Architectures \label{sec: model-architectures}}
\vspace{-0.2cm}

\subsubsection{Generating Neural Observations Conditioned on Codes with High Information \label{sec: info-diff-methods}}
Our model is an adaptation of InfoDiffusion \cite{InfoDiffusion} that operates on time-series data. The model consists of two networks: the auxiliary variable encoder, and the noise prediction network. The architecture of both the auxiliary encoder and the noise predictor is a bidirectional GRU with a linear readout to the desired output dimensionality. The auxiliary variable encoder is passed a window of neural activity $\{\textbf{n}_t\}_{t=1}^T \in \mathbb{R}^{T \times N}$ as input, and outputs the latent representation vector, or code, $\textbf{c} \in \mathbb{R}^{L}$. The code is penalized against a standard normal prior using Maximum Mean Discrepancy (MMD) \cite{MMD}. This code $\textbf{c}$ is used to condition a diffusion process, wherein the noise prediction network is trained to predict the noise that is added to the data during the forward process for any step sampled randomly from the sequence of steps in the total diffusion process, consistent with the score matching objective and training procedure described in \cite{DDPM}. The forward process is defined by Equation \ref{forward_process}:
\begin{align}
    \tilde{\textbf{x}}_{i} = \sqrt{\bar{\alpha}_i}\textbf{x}_0 + \sqrt{1-\bar{\alpha}_i}\epsilon \ ; \  \epsilon \sim \mathcal{N}(0,1) \label{forward_process}
\end{align}
Where $\textbf{x}_0$ is a window of neural activity, $\tilde{\textbf{x}}_{i}$ is the noised neural data sample at step $i$ of the forward process, $\bar{\alpha}_i$ is the cumulative product of the coefficients $\alpha_i = 1-\beta_i$ up to diffusion step $i$, where the parameters $\beta_i \in [0.001, 0.01]$ are fixed and define a linear noising schedule of the forward process.
\begin{figure}[h]
  \centering
  \includegraphics[width=0.8\textwidth]{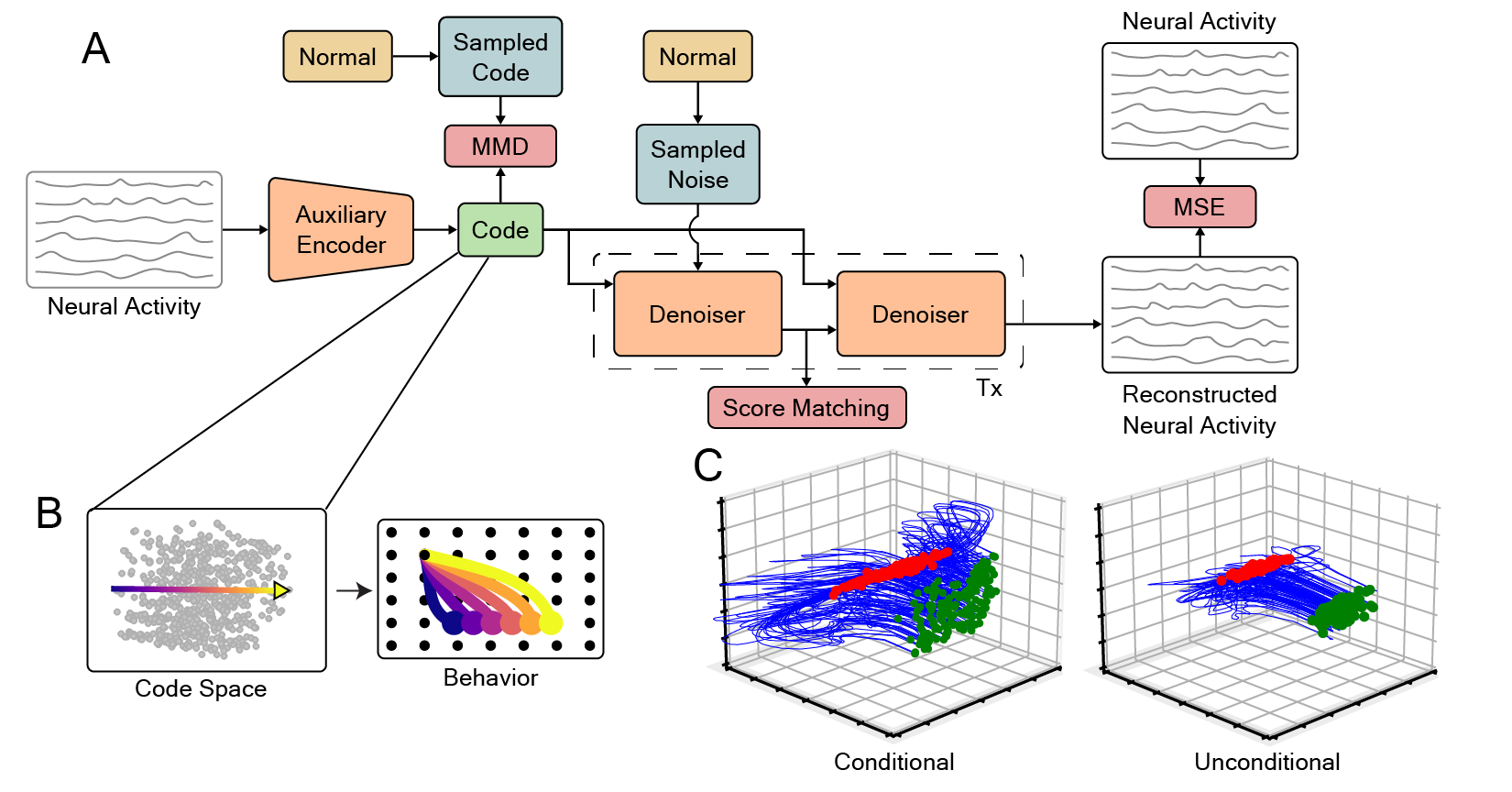}
  \caption{\textbf{GNOCCHI model overview}. A) Codes are inferred from neural activity and used for conditional generation.  Using the noise predicted by the denoiser, the neural activity is reconstructed and penalized against the original neural activity (Mean Squared Error). During inference, Gaussian white noise is iteratively transformed into a sample using the inferred code and the denoiser. B) A schematic of how a movement in code space corresponds to a change in target position. C) Low-dimensional visualizations of the generated neural activity during conditional (left) and unconditional (right) generation.}
  \label{fig:GNOCCHI model schematic}
  \vspace{-0.2cm}
\end{figure}
In addition to the MMD and score matching costs, there is a cost on the fidelity of the reconstructed neural data sample $\hat{\textbf{x}}_0$ computed using the output of the noise predictor $\tilde{\epsilon}$ and the noised neural data sample  $\tilde{\textbf{x}_i}$. In total, the relevant quantities for training the model are given by Equations \ref{compute_codes}-\ref{reconstruct_data_sample} :
\begin{align}
    \textbf{c} &= \text{AuxEncoder}(\{\textbf{n}_t\}_{t=1}^T) \label{compute_codes}\\
    \tilde{\epsilon}_i &= \text{NoisePredictor}(\tilde{\textbf{x}}_i, i, \textbf{c}) \label{predict_noise_added} \\
    \hat{\textbf{x}}_0 &= \frac{1}{\sqrt{\bar{\alpha}_i}}\left(\tilde{\textbf{x}_i} - \sqrt{1-\bar{\alpha}_i}\tilde{\epsilon}_i\right) \label{reconstruct_data_sample}
\end{align}
Note that Equation \ref{reconstruct_data_sample} is simply a rearrangement of Equation \ref{forward_process} with the reconstructed neural data sample $\hat{\textbf{x}}_0$ used instead of the original neural data sample $\textbf{x}_0$ and the predicted noise $\tilde{\epsilon}$ used in place of the sampled noise $\epsilon$. In total, the loss terms computed based on these quantities are given by Equations \ref{score_loss}-\ref{MMD_loss}:
\begin{align}
    \mathcal{L}_{\text{score}} &= \mathbb{E}\left[||\epsilon_i - \tilde{\epsilon}_i||_2^2\right] \label{score_loss}\\
    \mathcal{L}_{\text{recon}} &=  \mathbb{E}\left[||\textbf{x}_0 - \hat{\textbf{x}}_0||_2^2\right] \label{recon_loss} \\
    \mathcal{L}_{\text{MMD}} = \mathbb{E}_{\textbf{c}_{\text{prior}} \sim \mathcal{N}(0,1)}\left[k(\textbf{c}_{\text{prior}},\textbf{c}_{\text{prior}})\right] &+ \mathbb{E}_{\textbf{c} \sim \text{AuxEnc}}\left[k(\textbf{c},\textbf{c}) \right] - 2\mathbb{E}_{\textbf{c}_{\text{prior} \sim \mathcal{N}(0,1)}, \textbf{c} \sim \text{AuxEnc}}\left[k(\textbf{c}_{\text{prior}},\textbf{c}) \right] \label{MMD_loss}
\end{align}
Where $k(\cdot, \cdot) : \mathbb{R}^L \times \mathbb{R}^L \rightarrow \mathbb{R}$ is the Gaussian radial basis function kernel.

In total, the loss function is the weighted sum of $\mathcal{L}_{\text{MMD}}, \mathcal{L}_{\text{score}}$ and $\mathcal{L}_{\text{recon}}$. The weightings of these terms as well as the rest of the hyperparameters used for the experiments are given in Section \ref{appendix: GNOCCHI}. The weighting and implementation of the MMD term is computed analogously to the InfoVAE implementation of \cite{PyTorch-VAE}. Additional details on the model architecture and hyperparameter selection are also in Section \ref{appendix: GNOCCHI}. A model schematic showing the training and inference pipelines for GNOCCHI is shown in Figure \ref{fig:GNOCCHI model schematic}, as well as low-dimensional visualizations of generated neural activity when conditioning on the codes, versus sampling from the code prior (Fig. \ref{fig:GNOCCHI model schematic}C).

\subsection{Latent Factor Analysis via Dynamical Systems (LFADS) \label{sec: lfads-models}}

LFADS\cite{LFADS_original, lfads-torch} is a sequential adaptation of the variational autoencoder (VAE) \cite{AEVB} that incorporates inductive biases from the dynamical systems perspective \cite{dynamical_systems_perspective} into an end-to-end latent dynamics model of neural activity. LFADS utilizes recurrent neural networks (RNNs) to read over neural activity and produce a summarizing vector representation, which is used as the initial condition of a dynamical system that is learned and unrolled by another RNN. The resulting architecture is capable of learning variational parameters that can be used to encode a dynamics-based representation of the neural recordings, expressed using a lower-dimensional set of latent variables. We trained LFADS models with the initial condition representation matched to the same dimensionality as the representations used with GNOCCHI. Additionally, we turned off the LFADS controller, since the appropriate strategy for taking time-varying inputs into account when performing latent navigation (see Section \ref{sec: latent-navigation}) is unclear. More details on the training, architecture choices and hyperparameters used with LFADS can be found in Section \ref{appendix: lfads}. \\

\subsection{Generating neural activity reflecting unseen conditions via Latent Navigation \label{sec: latent-navigation}}
\vspace{-0.2cm}
To generate condition-specific samples from the latent space learned by the generative models, we developed a simple method to find locations in the latent space that corresponded to trials with specific behavioral features (e.g., for the random target tasks, starting location and desired target location.) This method, termed \textit{latent navigation}, consists of fitting a linear regression from the latent representations learned by the generative model, $\textbf{c} \in \mathbb{R}^{L}$ to each behavioral feature. We then obtain new samples related to each behavioral feature by using latents obtained by stepping along the directions learned by the linear mapping, starting from a reference point in the latent space, referred to as an ``anchor latent". This process is summarized by Equations \ref{lin_map_for_interp} and \ref{new_codes_from_interp}:
\begin{align}
    &\min_{\textbf{W}, \textbf{b}}||\textbf{s} - \textbf{Wc} + \textbf{b}||^2_2 \label{lin_map_for_interp}\\ 
    \tilde{\textbf{c}}^{(i)}_j &= \textbf{c}_{anchor} \pm i \cdot \delta c \cdot \textbf{w}_j \label{new_codes_from_interp}
\end{align}
Where $\textbf{s} \in \mathbb{R}^{K}$ is the set of behavioral features, $\textbf{W} \in \mathbb{R}^{K\times L}$ is the matrix fit by linear regression with bias $\textbf{b} \in \mathbb{R}^{K}$, $\textbf{w}_j$ is the $j^{th}$ row of $\textbf{W}$, $i$ indicates how many steps are being taken away from the anchor latent, and $\delta c \in \mathbb{R}$ is the step size in latent space. In this work, we considered $j$ corresponding to the horizontal and vertical target coordinates. For the artificial random target task, we used $\delta c = 1.0$.

\subsection{Behavioral Decoding \label{sec: behavioral-decoding}}
\vspace{-0.2cm}
To assess the nature of behavior that can be produced by conditionally generating neural activity, we decode behavior from the generated neural activity for both seen and unseen conditions. For the artificial neural datasets, we predicted the hand endpoint position. 
The position decoder $f : \mathbb{R}^{N} \rightarrow \mathbb{R}^{2}$ is given by Equation \ref{artificial-neural-decoder}:
\begin{align}
    \textbf{x}_t &= f(\textbf{n}_t) \label{artificial-neural-decoder}
\end{align}
Where $\textbf{x}_t$ is the endpoint position at time $t$, $\textbf{n}_t$ is the neural activity at time $t$. For $f$, we use ridge regression \cite{scikit-learn}.

The position decoders for the artificial neural response experiments are fit on the ground truth neural responses and behaviors, and we use these decoders to assess the responses produced by the generative models. The ridge regularization parameter of the position decoders are set to $1.0$ to mitigate noise differences between the model-produced neural activity and the ground truth RNN hidden states used to train the decoders.

\section{Results}
\vspace{-0.2cm}
\subsection{GNOCCHI learns well-structured codes that reflect task variables}
We  aimed to validate three features of the GNOCCHI model: 1) unsupervised discovery of structured codes, 2) that those codes should be disentangled, and 3) that the model would generate more accurate samples for unseen behavioral conditions than alternative models. 

\begin{figure*}[h]
    \centering
    \includegraphics[width = \textwidth]{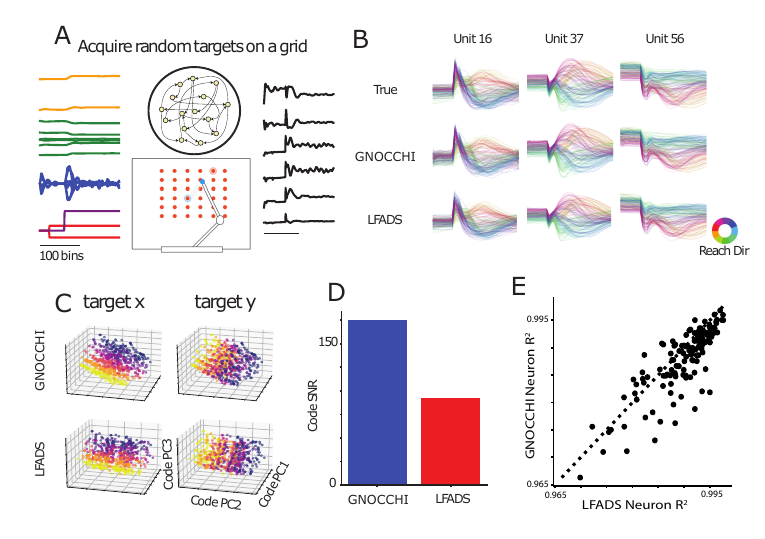}
    \vspace{-6mm}
    \caption{\textbf{Validating GNOCCHI on realistic synthetic neural activity}. A) RNN task training overview. We trained an RNN to control a biomechanical effector to manipulate the endpoint to acquire targets on a grid. The RNN received task inputs indicating the target location and the go cue time, as well as sensory feedback inputs related the effector endpoint, as well as muscle lengths and velocities. The RNN produced activations of a set of muscles to control the arm based on these inputs. B) Visualizing neural responses colored by the relative angle of the reach. Both generative models produced unit responses that closely resemble both the timecourse and behavioral structure of the ground truth data. C) Top 3 principal components of the codes. Codes exhibited organization according to several behavioral variables;  here they are colored by the target x and y locations of individual trials. D) Code signal-to-noise ratio. The codes learned by GNOCCHI had a substantially higher SNR for target position than those learned by LFADS, indicating a closer relationship to behavior. E) Additionally, computing the $R^2$ between the ground truth and generated unit activity quantifies that GNOCCHI-generated activity matched the ground truth with comparable accuracy to LFADS.}
    \label{fig:SyntheticFull}
    \vspace{-0.2cm}
\end{figure*}

We began by applying GNOCCHI to a complex, biologically-plausible synthetic dataset with known ground-truth activity. We generated this dataset by training an RNN to perform a Random Target reaching task by controlling a 2 DoF manipulandum in a simulation environment (Fig. \ref{fig:SyntheticFull}A). After training, individual RNN units had complex, time-varying activity patterns that were well-modulated by task parameters (Fig \ref{fig:SyntheticFull}B, top row, traces colored by angle of reach) and qualitatively resembled typical response patterns of motor cortical neurons. We fit GNOCCHI and LFADS models to this synthetic neural activity and found both models were able to accurately reconstruct single unit activities (Fig. \ref{fig:SyntheticFull}B, lower rows).

We then examined the latent representations inferred by each model and found that while both sets of latents had clear structure related to the target position, GNOCCHI’s codes appeared more tightly organized than LFADS (Fig. \ref{fig:SyntheticFull}C). To quantify how tightly the codes were clustered for a particular reach condition, we computed a signal-to-noise ratio (SNR) metric, the ratio of the total variance to the mean intra-condition variance. The SNR of GNOCCHI codes was substantially higher than for LFADS (Fig. \ref{fig:SyntheticFull}D), suggesting that GNOCCHI learned more structured code representations that better reflected the underlying behavior.

\subsection{GNOCCHI learns disentangled codes that allow for targeted data generation}

\begin{figure*}[h]
    \centering
    \includegraphics[width = 0.95\textwidth]{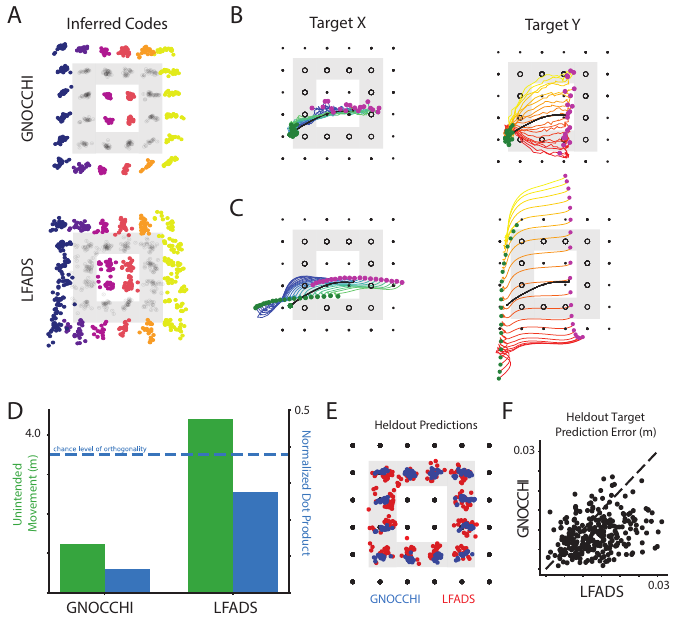}
    \vspace{-3mm}
    \caption{ \textbf{Controllable generation of neural activity for novel behavioral conditions}. A) Target-tuned subspaces. Each model contained code subspaces with a close correspondence to target location. Codes for the held-in conditions are colored by the target x location, and the held-out codes are black. Code organization matched the target grid for seen and unseen conditions for both models. B, C) Behavior decoded from trials produced by latent navigation. When we generate new samples by navigating along the target x (left) and target y (right) dimensions of code space, the trajectories produced by GNOCCHI vary in the direction of interest and largely isolate the intended variable (endpoint location), while for LFADS there is coupling with other variables such as the start location of the reach. D) Unintended movement and the orthogonality of behavioral dimensions in code space. We quantified unwanted variation in the decoded behavior as the total absolute deviation of each of the variables that should remain fixed during latent navigation of a single variable (green). Consistent with the decoding visualization (B and C), unwanted movements produced by LFADS were substantially higher than those produced by GNOCCHI. Additionally, we compute the normalized dot product between the vectors that define each behavioral direction during latent navigation (blue), and found that the directions in the code space of LFADS are less orthogonal / more strongly coupled than for GNOCCHI. E, F) Heldout predicted target position from inferred codes. Passing the heldout trials through each trained model, we find that both GNOCCHI and LFADS are capable of representing trials with structure that generalizes well to the task.}
    \label{fig:SyntheticHoldout}
    \vspace{-0.2cm}
\end{figure*}

A disentangled code space allows a generative model to produce new samples in which the relevant task variables encoded by the data change independently from one another. To test whether GNOCCHI learns a disentangled code space, we trained GNOCCHI and LFADS models on a subset of the previous Random Target task data, only including conditions in the middle or on the edges of the workspace (Fig. \ref{fig:SyntheticHoldout}A).

Using our latent navigation procedure, we isolated the “Target-tuned” axes of code space, generated artificial neural data with codes associated with changing a single task-variable, and used a position decoder to transform these artificial neural signals into a predicted hand position over the course of a trial. When we generated activity by moving along the target-tuned dimensions of the latent space learned by GNOCCHI, the decoded position varied for the behavioral variable of interest largely in isolation, e.g. with minimal change in the start position when varying along the target location directions of code space (Fig \ref{fig:SyntheticHoldout}B, left, right, respectively). 

In contrast, moving along both the X and Y target-tuned dimensions of the LFADS latent space resulted in generated activity for which the decoded start position varied substantially for both target dimensions (Fig. \ref{fig:SyntheticHoldout}C, left: Target X, right: Target Y), indicating an entangled representation for start and target locations in the LFADS model. Additionally, interpolating novel samples within the range of heldout targets resulted in decoded trajectories that ended far outside the intended boundary for LFADS, while decoding samples that were generated through GNOCCHI produced trajectories that were largely confined to the intended range.

We quantified these observations in two ways. First, we indirectly measured disentanglement by calculating the pair-wise normalized dot products between the axes of the linear mapping used for latent navigation. When these values are close to zero, the learned representations are nearly orthogonal in the code space. The average projection magnitude was significantly larger for LFADS compared to GNOCCHI (Fig. \ref{fig:SyntheticHoldout}D, paired t-test across 10-fold cross validation,.p-val < 2.8e-34). For reference, we include a horizontal line indicating the average magnitude of the normalized dot product for vectors of the same dimension as the target-tuned axes, whose entries were drawn from a standard normal distribution. Second, we calculated the effect of moving along one target-tuned dimension on the other variables. Using our decoders, we quantified how much unintended movement in non-queried variables was observed as we changed the target-tuned code. We found that the unintended movement was substantially larger for LFADS than for GNOCCHI (Fig. \ref{fig:SyntheticHoldout}D, paired t-test across 10 different reach interpolations,.p-val < 8.6e-14). Together, these results suggest that GNOCCHI learns a more disentangled representation of task variables than LFADS.

Last, a desired feature of generative models is to obtain high quality generated samples both within the domain of the training and to generalize outside of the training set. To test the ability of our models to generalize, we used neural activity from the held-out conditions to infer codes with the LFADS and GNOCCHI code encoders (Fig. \ref{fig:SyntheticHoldout}E). We found that the codes inferred by GNOCCHI were generally closer to the correct location in code space, suggesting that GNOCCHI can generate higher quality samples from unseen conditions than the alternative model (Fig. \ref{fig:SyntheticHoldout}F).

\subsection{Biological Neural Data}
\begin{figure*}[h]
    \centering
    \includegraphics[width = \textwidth]{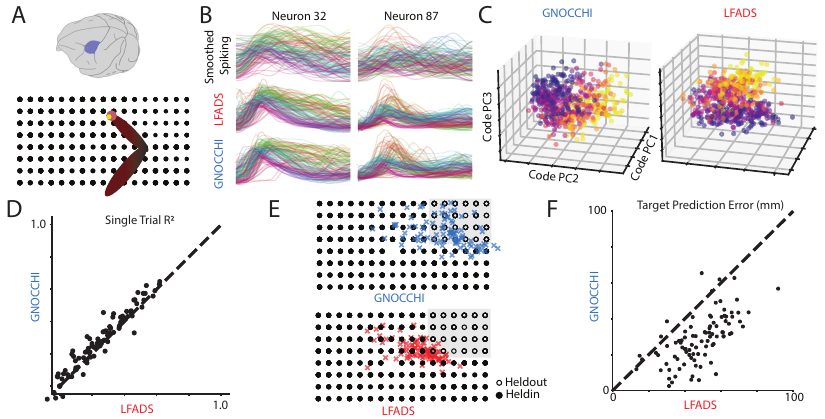}
    \vspace{-6mm}
    \caption{\textbf{GNOCCHI generates realistic neural activity and captures held-out conditions in a biological dataset}. A) Task schematic. Recordings from primary motor cortex of a monkey (top) while performing a random-target reaching task (bottom). B) Single neuron responses from two example neurons (columns) in the window [-200, 500] aligned to movement onset, colored by reach direction. Top: Smoothed single trial activity. Middle, Bottom: Predicted single trial activity from LFADS/GNOCCHI, respectively. C) Visualization of top 3 PCs of the code space for GNOCCHI (left) and LFADS (right), color coded by Target X position. D) $R^2$ for predicted activity of individual neurons across validation trials for LFADS (x-axis) and GNOCCHI (y-axis). Points above the unity line denote neurons in which GNOCCHI predicted activity that more closely resembled the smoothed firing rates. E) Diagram of heldout generalization experiment. Heldin (heldout) trial target locations indicated with filled (unfilled) circles. Predicted target location from inferred codes indicated by colored $\times$s (LFADS: red, GNOCCHI: blue). F) Scatter plot of error between actual target location and predicted target location (from E) for GNOCCHI (x-axis) and LFADS (y-axis). Points below the unity line indicate trials where the GNOCCHI prediction was closer to the true target than LFADS.}
    \label{fig:NeuralData}
\end{figure*}

We applied GNOCCHI and LFADS to neural data from the motor cortex of a monkey performing a similar random-target reaching task as our synthetic dataset (Fig. \ref{fig:NeuralData}A). Both models were able to capture significant features of single trial neural responses (Fig. \ref{fig:NeuralData}B, single trial neural activity color-coded by reach angle) and inferred codes that had significant structure that reflected the underlying task (Fig. \ref{fig:NeuralData}C, codes color-coded by Target X position). 

We first wanted to compare the ability of GNOCCHI to model the neural activity compared to the LFADS; higher predictive performance would reflect a more structured code space. We compared the $R^2$ values of LFADS to GNOCCHI, and found that on average GNOCCHI made better predictions of the smoothed single trial neural activity across neurons (Fig \ref{fig:NeuralData}D, paired t-test across neurons,.p-val < 1e-4). We also tested whether the representations learned by GNOCCHI were disentangled, using the same methods as applied to the synthetic data (Fig \ref{fig:NeuralData}B, E). 

We next tested the ability of GNOCCHI to generalize to unseen conditions. We trained new GNOCCHI and LFADS models on a subset of target locations, leaving a quadrant of the target locations out of the training set (Fig \ref{fig:NeuralData}E, Heldin, filled circles, Heldout unfilled circles). We then used the neural activity from the heldout conditions to infer codes for each model. When we passed the inferred codes through a decoder trained to predict held-in target position from heldin codes, we found that GNOCCHI codes predicted reaches that extended much further into the heldout targets than did the LFADS codes (Fig \ref{fig:NeuralData}E, blue (top), red (bottom)  $\times$s, respectively).  We quantified this boost in generalization performance by the distance between the target predicted by the code to the true target of each trial. We found that GNOCCHI codes were more predictive of the true target than LFADS for almost all of the heldout trials (Fig \ref{fig:NeuralData}F, points below unity line were closer for GNOCCHI than LFADS), demonstrating the utility of GNOCCHI for generating out-of-training neural activity relative to LFADS.

\section{Limitations}
\vspace{-0.2cm}
Aside from some of the comparative advantages and disadvantages of each model, there are some limitations to GNOCCHI as proposed here. One limitation of this work is that the models we considered only generate neural activity, and not also behavior. Without generating behavior and neural activity concurrently, we cannot augment datasets for training neural decoders without making additional assumptions on how the generated neural responses relate to behavior. However, GNOCCHI may be appropriate for decoding applications (e.g., brain-computer interfaces; BCIs) as-is. Typically, in clinical applications with paralyzed individuals, behavior is not directly observed; rather, decoders are trained by making assumptions about the relationship between neural activity and behavior, and those same assumptions could be applied to samples generated by GNOCCHI. However, integrating GNOCCHI with methods that can jointly generate neural activity and behavior from a shared latent representation \cite{JointNeuroBehavior} presents a promising next step for this work, though the matter of overcoming the lack of ground truth behavior in clinical settings would remain a challenge. Additionally, the data in this work are aligned to a stimulus (go cue) or movement onset. Achieving interpretable representations that support latent navigation when modeling continuous time-series data (without pre-specified alignment) would enable application to a broader range of behaviors, and validating the efficacy of GNOCCHI in such settings is an important improvement for future work.

This work demonstrates an exciting opportunity for increasing both the generative capability and representational interpretability of neural population models. Such improvements could enable higher-performing BCIs, which are used in rehabilitation settings to enable increased autonomy and quality of life for those with neurological conditions. Given the potential impact of this work on BCI, we encourage appropriate consideration to preserving personal privacy and equity in any such downstream applications.

\section{Discussion}
\vspace{-0.2cm}
GNOCCHI leverages advances from InfoDiffusion to learn structured, disentangled, and generalizable code spaces without requiring behavioral labels. GNOCCHI builds upon previous advances in generative modeling, which have provided powerful new lenses into neural population activity. Such advances will be critical as neuroscientists gain access to high-dimensional neural population recordings from a variety of brain areas during increasingly complex and unconstrained behaviors. The ability of methods like GNOCCHI to interpolate or even extrapolate neural data samples along interpretable latent axes may provide the opportunity to probe high dimensional behaviors with greater sampling efficiency, leading to a deeper understanding of the structure and geometry of neural activity during complex behaviors.

\begin{ack}
\vspace{-0.2cm}
This work was supported by NIH-NINDS/OD DP2NS127291, NIH BRAIN/NIDA RF1 DA055667, and the Simons Foundation as part of the Simons-Emory International Consortium on Motor Control (CP). This work was also supported by NIH BRAIN Initiative grant 5F32MH132175 (CV) and NIH NIBIB T32EB025816 (DM). 

CP is a consultant for Meta (Reality Labs). This entity did not support this work, did not have a role in the study, and did not have any competing interests related to this work. 

The authors would like to acknowledge Brianna Karpowicz for discussions that helped shape the analysis and pre-processing of the biological neural population data.

\end{ack}

\printbibliography

@misc{InfoVAE,
      title={InfoVAE: Information Maximizing Variational Autoencoders}, 
      author={Shengjia Zhao and Jiaming Song and Stefano Ermon},
      year={2018},
      eprint={1706.02262},
      archivePrefix={arXiv},
      primaryClass={cs.LG}
}

@InProceedings{InfoDiffusion,
  title = 	 {{I}nfo{D}iffusion: Representation Learning Using Information Maximizing Diffusion Models},
  author =       {Wang, Yingheng and Schiff, Yair and Gokaslan, Aaron and Pan, Weishen and Wang, Fei and De Sa, Christopher and Kuleshov, Volodymyr},
  booktitle = 	 {Proceedings of the 40th International Conference on Machine Learning},
  pages = 	 {36336--36354},
  year = 	 {2023},
  editor = 	 {Krause, Andreas and Brunskill, Emma and Cho, Kyunghyun and Engelhardt, Barbara and Sabato, Sivan and Scarlett, Jonathan},
  volume = 	 {202},
  series = 	 {Proceedings of Machine Learning Research},
  month = 	 {23--29 Jul},
  publisher =    {PMLR},
  url = 	 {https://proceedings.mlr.press/v202/wang23ah.html}
}

@misc{LFADS_original,
      title={LFADS - Latent Factor Analysis via Dynamical Systems}, 
      author={David Sussillo and Rafal Jozefowicz and L. F. Abbott and Chethan Pandarinath},
      year={2016},
      eprint={1608.06315},
      archivePrefix={arXiv},
      primaryClass={cs.LG}
}

@misc{AEVB,
      title={Auto-Encoding Variational Bayes}, 
      author={Diederik P Kingma and Max Welling},
      year={2022},
      eprint={1312.6114},
      archivePrefix={arXiv},
      primaryClass={stat.ML}
}

@article{dynamical_systems_perspective,
  title={Cortical control of arm movements: a dynamical systems perspective.},
  author={Krishna V. Shenoy and Maneesh Sahani and Mark M. Churchland},
  journal={Annual review of neuroscience},
  year={2013},
  volume={36},
  pages={
          337-59
        },
}

@inproceedings{piVAE,
 author = {Zhou, Ding and Wei, Xue-Xin},
 booktitle = {Advances in Neural Information Processing Systems},
 editor = {H. Larochelle and M. Ranzato and R. Hadsell and M.F. Balcan and H. Lin},
 pages = {7234--7247},
 publisher = {Curran Associates, Inc.},
 title = {Learning identifiable and interpretable latent models of high-dimensional neural activity using pi-VAE},
 url = {https://proceedings.neurips.cc/paper_files/paper/2020/file/510f2318f324cf07fce24c3a4b89c771-Paper.pdf},
 volume = {33},
 year = {2020}
}

@dataset{ODoherty_RTT_2017,
  author       = {O'Doherty, Joseph E. and
                  Cardoso, Mariana M. B. and
                  Makin, Joseph G. and
                  Sabes, Philip N.},
  title        = {{Nonhuman Primate Reaching with Multichannel 
                   Sensorimotor Cortex Electrophysiology}},
  month        = may,
  year         = 2017,
  %note         = {{This research was supported by the Congressionally 
     %              Directed Medical Research Program
      %             (W81XWH-14-1-0510). JEO was supported by
       %            fellowship \#2978 from the Paralyzed Veterans of
        %           America. JGM was supported by a fellowship from
         %          the Swartz Foundation.}},
  publisher    = {Zenodo},
  doi          = {10.5281/zenodo.583331},
  %url          = {https://doi.org/10.5281/zenodo.583331}
}

@inproceedings{
SwapVAE,
title={Drop, Swap, and Generate: A Self-Supervised Approach for Generating Neural Activity},
author={Ran Liu and Mehdi Azabou and Max Dabagia and Chi-Heng Lin and Mohammad Gheshlaghi Azar and Keith B Hengen and Michal Valko and Eva L Dyer},
booktitle={Advances in Neural Information Processing Systems},
editor={A. Beygelzimer and Y. Dauphin and P. Liang and J. Wortman Vaughan},
year={2021},
url={https://openreview.net/forum?id=ZRPRjfAF3yd}
}

@article{LFADS_NatMethods,
  title={Inferring single-trial neural population dynamics using sequential auto-encoders},
  author={Chethan Pandarinath and Daniel J. O’Shea and Jasmine Collins and Rafal J{\'o}zefowicz and Sergey D. Stavisky and Jonathan C. Kao and Eric M. Trautmann and Matthew T. Kaufman and Stephen I. Ryu and Leigh R. Hochberg and Jaimie M. Henderson and Krishna V. Shenoy and L. F. Abbott and David Sussillo},
  journal={Nature methods},
  year={2017},
  volume={15},
  pages={805 - 815},
}

@article{MotorNet, title={MotorNet: a Python toolbox for controlling differentiable biomechanical effectors with artificial neural networks}, url={http://dx.doi.org/10.1101/2023.02.17.528969}, DOI={10.1101/2023.02.17.528969}, publisher={Cold Spring Harbor Laboratory}, author={Codol, Olivier and Michaels, Jonathan A. and Kashefi, Mehrdad and Pruszynski, J. Andrew and Gribble, Paul L.}, year={2023}, month=feb }

@misc{lfads-torch,
      title={lfads-torch: A modular and extensible implementation of latent factor analysis via dynamical systems}, 
      author={Andrew R. Sedler and Chethan Pandarinath},
      year={2023},
      eprint={2309.01230},
      archivePrefix={arXiv},
      primaryClass={cs.LG}
}

@article{recording_affects_analysis,
  title={How advances in neural recording affect data analysis},
  author={Ian H. Stevenson and Konrad Paul Kording},
  journal={Nature Neuroscience},
  year={2011},
  volume={14},
  pages={139-142},
}

@article{Neuropixels,
author = {Nicholas A. Steinmetz  and Cagatay Aydin  and Anna Lebedeva  and Michael Okun  and Marius Pachitariu  and Marius Bauza  and Maxime Beau  and Jai Bhagat  and Claudia Böhm  and Martijn Broux  and Susu Chen  and Jennifer Colonell  and Richard J. Gardner  and Bill Karsh  and Fabian Kloosterman  and Dimitar Kostadinov  and Carolina Mora-Lopez  and John O’Callaghan  and Junchol Park  and Jan Putzeys  and Britton Sauerbrei  and Rik J. J. van Daal  and Abraham Z. Vollan  and Shiwei Wang  and Marleen Welkenhuysen  and Zhiwen Ye  and Joshua T. Dudman  and Barundeb Dutta  and Adam W. Hantman  and Kenneth D. Harris  and Albert K. Lee  and Edvard I. Moser  and John O’Keefe  and Alfonso Renart  and Karel Svoboda  and Michael Häusser  and Sebastian Haesler  and Matteo Carandini  and Timothy D. Harris },
title = {Neuropixels 2.0: A miniaturized high-density probe for stable, long-term brain recordings},
journal = {Science},
volume = {372},
number = {6539},
pages = {eabf4588},
year = {2021},
doi = {10.1126/science.abf4588},
URL = {https://www.science.org/doi/abs/10.1126/science.abf4588},
eprint = {https://www.science.org/doi/pdf/10.1126/science.abf4588}}

@article{scikit-learn,
  title={Scikit-learn: Machine Learning in {P}ython},
  author={Pedregosa, F. and Varoquaux, G. and Gramfort, A. and Michel, V.
          and Thirion, B. and Grisel, O. and Blondel, M. and Prettenhofer, P.
          and Weiss, R. and Dubourg, V. and Vanderplas, J. and Passos, A. and
          Cournapeau, D. and Brucher, M. and Perrot, M. and Duchesnay, E.},
  journal={Journal of Machine Learning Research},
  volume={12},
  pages={2825--2830},
  year={2011}
}

@article{UtahArray,
title = {The Utah Intracortical Electrode Array: A recording structure for potential brain-computer interfaces},
journal = {Electroencephalography and Clinical Neurophysiology},
volume = {102},
number = {3},
pages = {228-239},
year = {1997},
issn = {0013-4694},
doi = {https://doi.org/10.1016/S0013-4694(96)95176-0},
url = {https://www.sciencedirect.com/science/article/pii/S0013469496951760},
author = {Edwin M. Maynard and Craig T. Nordhausen and Richard A. Normann}
}

@article{sedler2022expressive,
  title={Expressive architectures enhance interpretability of dynamics-based neural population models},
  author={Sedler, Andrew R and Versteeg, Christopher and Pandarinath, Chethan},
  journal={Neurons, Behavior, Data analysis, and Theory},
  year={2023}
}

@inproceedings{
ODIN,
title={Expressive dynamics models with nonlinear injective readouts enable reliable recovery of latent features from neural activity},
author={Christopher Versteeg and Andrew Sedler and Jonathan McCart and Chethan Pandarinath},
booktitle={NeurIPS 2023 Workshop on Symmetry and Geometry in Neural Representations},
year={2023},
url={https://openreview.net/forum?id=kLwwaBdWAJ}
}

@article{vetter2023generating,
  title={Generating realistic neurophysiological time series with denoising diffusion probabilistic models},
  author={Vetter, Julius and Macke, Jakob H and Gao, Richard},
  journal={bioRxiv},
  pages={2023--08},
  year={2023},
  publisher={Cold Spring Harbor Laboratory}
}

@article{zhu2022deep,
  title={A deep learning framework for inference of single-trial neural population dynamics from calcium imaging with subframe temporal resolution},
  author={Zhu, Feng and Grier, Harrison A and Tandon, Raghav and Cai, Changjia and Agarwal, Anjali and Giovannucci, Andrea and Kaufman, Matthew T and Pandarinath, Chethan},
  journal={Nature neuroscience},
  volume={25},
  number={12},
  pages={1724--1734},
  year={2022},
  publisher={Nature Publishing Group US New York}
}

@inproceedings{
beta-VAE,
title={beta-{VAE}: Learning Basic Visual Concepts with a Constrained Variational Framework},
author={Irina Higgins and Loic Matthey and Arka Pal and Christopher Burgess and Xavier Glorot and Matthew Botvinick and Shakir Mohamed and Alexander Lerchner},
booktitle={International Conference on Learning Representations},
year={2017},
url={https://openreview.net/forum?id=Sy2fzU9gl}
}

@article{MMD,
  author  = {Arthur Gretton and Karsten M. Borgwardt and Malte J. Rasch and Bernhard Sch{{\"o}}lkopf and Alexander Smola},
  title   = {A Kernel Two-Sample Test},
  journal = {Journal of Machine Learning Research},
  year    = {2012},
  volume  = {13},
  number  = {25},
  pages   = {723-773},
  url     = {http://jmlr.org/papers/v13/gretton12a.html}
}

@inproceedings{DDPM,
 author = {Ho, Jonathan and Jain, Ajay and Abbeel, Pieter},
 booktitle = {Advances in Neural Information Processing Systems},
 editor = {H. Larochelle and M. Ranzato and R. Hadsell and M.F. Balcan and H. Lin},
 pages = {6840--6851},
 publisher = {Curran Associates, Inc.},
 title = {Denoising Diffusion Probabilistic Models},
 url = {https://proceedings.neurips.cc/paper_files/paper/2020/file/4c5bcfec8584af0d967f1ab10179ca4b-Paper.pdf},
 volume = {33},
 year = {2020}
}

@misc{PyTorch,
      title={PyTorch: An Imperative Style, High-Performance Deep Learning Library}, 
      author={Adam Paszke and Sam Gross and Francisco Massa and Adam Lerer and James Bradbury and Gregory Chanan and Trevor Killeen and Zeming Lin and Natalia Gimelshein and Luca Antiga and Alban Desmaison and Andreas Köpf and Edward Yang and Zach DeVito and Martin Raison and Alykhan Tejani and Sasank Chilamkurthy and Benoit Steiner and Lu Fang and Junjie Bai and Soumith Chintala},
      year={2019},
      eprint={1912.01703},
      archivePrefix={arXiv},
      primaryClass={cs.LG}
}

@misc{ray_tune,
      title={Tune: A Research Platform for Distributed Model Selection and Training}, 
      author={Richard Liaw and Eric Liang and Robert Nishihara and Philipp Moritz and Joseph E. Gonzalez and Ion Stoica},
      year={2018},
      eprint={1807.05118},
      archivePrefix={arXiv},
      primaryClass={cs.LG}
}

@software{PyTorch_Lightning,
author = {Falcon, William and {The PyTorch Lightning team}},
doi = {10.5281/zenodo.3828935},
license = {Apache-2.0},
month = mar,
title = {{PyTorch Lightning}},
url = {https://github.com/Lightning-AI/lightning},
version = {1.4},
year = {2019}
}

@article{AutoLFADS,
  title={A large-scale neural network training framework for generalized estimation of single-trial population dynamics},
  author={Mohammad Reza Keshtkaran and Andrew R. Sedler and Raeed H. Chowdhury and Raghav Tandon and Diya Basrai and Sarah L. Nguyen and Hansem Sohn and Mehrdad Jazayeri and Lee E. Miller and Chethan Pandarinath},
  journal={Nature Methods},
  year={2022},
}

@misc{CoordinatedDropout,
      title={Enabling hyperparameter optimization in sequential autoencoders for spiking neural data}, 
      author={Mohammad Reza Keshtkaran and Chethan Pandarinath},
      year={2019},
      eprint={1908.07896},
      archivePrefix={arXiv},
      primaryClass={cs.LG}
}

@article{JointNeuroBehavior,
author = {Schulz, Auguste and Vetter, Julius and Gao, Richard and Morales, Daniel and Lobato-Rios, Victor and Ramdya, Pavan and Goncalves, Pedro and Macke, Jakob},
journal={bioRxiv},
year = {2024},
month = {04},
pages = {},
title = {Modeling conditional distributions of neural and behavioral data with masked variational autoencoders},
doi = {10.1101/2024.04.19.590082}
}

@misc{PyTorch-VAE,
  author = {Subramanian, A.K},
  title = {PyTorch-VAE},
  year = {2020},
  publisher = {GitHub},
  journal = {GitHub repository},
  howpublished = {\url{https://github.com/AntixK/PyTorch-VAE}}
}

@misc{CtD_Benchmark,
  author = {Christopher Versteeg and Chethan Pandarinath},
  title = {Computation-Through-Dynamics Benchmark},
  year = {2024},
  publisher = {GitHub},
  journal = {GitHub repository},
  howpublished = {\url{https://github.com/snel-repo/ComputationThruDynamicsBenchmark}}
}

@inproceedings{GPFA,
 author = {Yu, Byron M and Cunningham, John P and Santhanam, Gopal and Ryu, Stephen and Shenoy, Krishna V and Sahani, Maneesh},
 booktitle = {Advances in Neural Information Processing Systems},
 editor = {D. Koller and D. Schuurmans and Y. Bengio and L. Bottou},
 pages = {},
 publisher = {Curran Associates, Inc.},
 title = {Gaussian-process factor analysis for low-dimensional single-trial analysis of neural population activity},
 url = {https://proceedings.neurips.cc/paper_files/paper/2008/file/ad972f10e0800b49d76fed33a21f6698-Paper.pdf},
 volume = {21},
 year = {2008}
}

@article{zhao2017variational,
  title={Variational latent gaussian process for recovering single-trial dynamics from population spike trains},
  author={Zhao, Yuan and Park, Il Memming},
  journal={Neural computation},
  volume={29},
  number={5},
  pages={1293--1316},
  year={2017},
  publisher={MIT Press One Rogers Street, Cambridge, MA 02142-1209, USA journals-info~…}
}

@article{wu2017gaussian,
  title={Gaussian process based nonlinear latent structure discovery in multivariate spike train data},
  author={Wu, Anqi and Roy, Nicholas A and Keeley, Stephen and Pillow, Jonathan W},
  journal={Advances in neural information processing systems},
  volume={30},
  year={2017}
}

@inproceedings{GP_factor_models,
 author = {Keeley, Stephen and Aoi, Mikio and Yu, Yiyi and Smith, Spencer and Pillow, Jonathan W},
 booktitle = {Advances in Neural Information Processing Systems},
 editor = {H. Larochelle and M. Ranzato and R. Hadsell and M.F. Balcan and H. Lin},
 pages = {13795--13805},
 publisher = {Curran Associates, Inc.},
 title = {Identifying signal and noise structure in neural population activity with Gaussian process factor models},
 url = {https://proceedings.neurips.cc/paper_files/paper/2020/file/9eed867b73ab1eab60583c9d4a789b1b-Paper.pdf},
 volume = {33},
 year = {2020}
}

@article{gao2016linear,
  title={Linear dynamical neural population models through nonlinear embeddings},
  author={Gao, Yuanjun and Archer, Evan W and Paninski, Liam and Cunningham, John P},
  journal={Advances in neural information processing systems},
  volume={29},
  year={2016}
}

@InProceedings{PLNDE,
  title = 	 {Inferring Latent Dynamics Underlying Neural Population Activity via Neural Differential Equations},
  author =       {Kim, Timothy D. and Luo, Thomas Z. and Pillow, Jonathan W. and Brody, Carlos D.},
  booktitle = 	 {Proceedings of the 38th International Conference on Machine Learning},
  pages = 	 {5551--5561},
  year = 	 {2021},
  editor = 	 {Meila, Marina and Zhang, Tong},
  volume = 	 {139},
  series = 	 {Proceedings of Machine Learning Research},
  month = 	 {18--24 Jul},
  publisher =    {PMLR},
  url = 	 {https://proceedings.mlr.press/v139/kim21h.html}
}

@article{schneider2023cebra,
  author={Schneider, Steffen and Lee, Jin Hwa and Mathis, Mackenzie Weygandt},
  title={Learnable latent embeddings for joint behavioural and neural analysis},
  journal={Nature},
  year={2023},
  month={May},
  day={03},
  issn={1476-4687},
  doi={10.1038/s41586-023-06031-6},
  url={https://doi.org/10.1038/s41586-023-06031-6}
}

@Misc{Yadan2019Hydra,
  author =       {Omry Yadan},
  title =        {Hydra - A framework for elegantly configuring complex applications},
  howpublished = {Github},
  year =         {2019},
  url =          {https://github.com/facebookresearch/hydra}
}

\appendix

\section*{Appendix}

\section{General dataset information}

\begin{table}[h]
  \caption{Full trial counts for each dataset}
 \label{tab:full_data-trial-counts}
  \centering
  \begin{tabular}{lll}
    \toprule
    Dataset     & Train     & Valid \\
    \midrule
    Random Target Task (artif.)     & 666 & 166      \\
    Random Target Task (bio.)     & 543       & 136  \\
    \bottomrule
  \end{tabular}
\end{table}

\begin{table}[h]
  \caption{Held-in and held-out trial counts for each dataset}
 \label{tab:heldin-heldout-trial-counts}
  \centering
  \begin{tabular}{llll}
    \toprule
    Dataset     & Train     & Valid & Heldout \\
    \midrule
    Random Target Task (artif.)     & 447 & 100 & 285    \\
    Random Target Task (bio.)     & 462  & 124 & 93  \\
    \bottomrule
  \end{tabular}
\end{table}

\section{Artificial neural activity from task-trained RNNs \label{appendix: task-trained}}

\subsection{Task training details}
The architecture of the RNN was a Gated Recurrent Unit (GRU). The RNNs were trained for 1000 epochs with learning rate $1\times 10^{-3}$ and no weight decay. The loss function was defined in terms of a weighted combination of the $L_1$ norm of the effector endpoint distance to the target, and an $L_2$ norm of the total activation of the muscles during behavior. Together with the variable timing of the go cue and a limited number of steps within a trial (see Section \ref{appendix: task-trained-rnn dataset sizes and trial info}), these loss terms incentivize timely completion of the task with minimal energy expenditure. The loss in total is given by Equation \ref{task_training_loss}:
\begin{align}
    \mathcal{L}_{\text{task}} = \frac{1}{2}\sum_i||\textbf{x}_i - \textbf{x}_{\text{target}}||_1 + \frac{1}{2}\sum_i||\textbf{a}_i||_2^2 \label{task_training_loss}
\end{align}
Where $\textbf{x} \in \mathbb{R}^2$ is the position of the effector endpoint, $\textbf{x}_{\text{target}} \in \mathbb{R}^2$ is the target location, and $\textbf{a} \in \mathbb{R}^6$ are the muscle activations. Additionally, a small ($10^{-3}$) amount of normally distributed noise was added to the outputs of the RNN to simulate motor noise.
\subsection{Data sizes, alignment, and trial information \label{appendix: task-trained-rnn dataset sizes and trial info}}
In each task, there were 1000 trials, split 800/200 between training and validation. The RNN has 128 hidden units, and each trial has a duration of 200 time steps. In order to analyze the neural activity of the reaching behavior, we aligned the data to a window of $[-10,30]$ time steps around the go cue. Since the go cue is given at a random time during the trial, any trials for which the go cue placement conflicted with the alignment window were discarded. The resulting number of trials for each task are given in Tables \ref{tab:full_data-trial-counts} and \ref{tab:heldin-heldout-trial-counts}. The resulting datasets used for modeling had trials consisting of 40 timesteps of activity across the 128 dimensions of the RNN during the window around the go cue.

\section{Generating Neural Observations Conditioned on Codes with High Information (GNOCCHI) \label{appendix: GNOCCHI}}

\subsection{Hyperparameter selection \label{appendix: GNOCCHI HP selection}}
To select hyperparameters for GNOCCHI, we searched for hyperparameters using a simplified task-trained RNN neural dataset, consisting of a single ring task in which the RNN produced reaches to a set of radially distributed targets starting from a common center point for each reach. We first ran a random search of 200 models, selecting hyper-parameters according to the ranges and sampling procedures given in Table \ref{tab:GNOCCHI_hp_ranges}, with final parameters given in Table \ref{tab:GNOCCHI_hps}. After this search, we selected the best model according to an even consideration of the MMD and score losses. To select the diffusion-based hyperparameters, we ran a grid search over the ranges given in Table \ref{tab:GNOCCHI_hps}, with all other parameters fixed. For each grid search, we also searched over 3 random seeds, and the parameter selected was the one for which the model performance was the best, with consideration given to the consistency across seeds. For the final hyperparameters, most of the choices were shared across all the datasets, only varying the number of diffusion steps or the latent dimensionality as needed (e.g. the code dimensionality for the random target task and biological random target task were 5 and 15 respectively, as shown by the bracketed values in Table \ref{tab:GNOCCHI_hps}). For the biological neural data, we analyzed the trends in MMD and score loss to assess whether the parameters chosen from the synthetic sweeps were effective. We found that increased code dimensionality and a larger number of diffusion steps led to models with better score loss, settling on 15 and 500 for the code dimensionality and number of diffusion steps respectively for modeling the biological neural data.

\begin{table}[h]
  \caption{Hyperparameter search: GNOCCHI}
 \label{tab:GNOCCHI_hp_ranges}
  \centering
  \begin{tabular}{llll}
    \toprule
    Parameter     & Search Type     & Search Range \\
    \midrule
    Latent Dimensionality & grid  & $[1,30]$ \\
    $\alpha_{\text{MMD}}$ & grid & $[-0.9,0]$ \\
    $\lambda_{\text{MMD}}$ & grid & $[1,1000]$\\
    score loss weight & grid & $[0.1,1]$ \\
    recon loss weight & grid & $[0.1,1]$ \\
    \# of diffusion steps & grid & $[25, 1500]$ \\
    $\beta$ range & grid & $[(0.00001,0.001), (0.01,0.06)]$ \\
    \midrule
    Auxiliary Encoder Hidden Size & random choice  & $[64, 128, 256, 512]$  \\
    Auxiliary Encoder Dropout  & uniform random & $[0.0,0.1]$    \\
    Auxiliary Encoder Learning Rate & random choice & $[0.000005, 0.00005, 0.0005]$  \\
    Auxiliary Encoder Weight Decay & random choice & $[0.0, 0.000001, 0.0001]$  \\
    Noise Predictor Hidden Size & random choice  & $[1024, 2048, 4096]$   \\
    Noise Predictor Dropout  & uniform random & $[0.0, 0.1]$ \\
    Noise Predictor Learning Rate & random choice & $[0.000005, 0.00005, 0.0005]$ \\
    Noise Predictor Weight Decay & random choice & $[0.0, 0.000001, 0.0001]$ \\
    Noise Predictor Positional Embedding Size & random choice & $[1, 5, 9, 13, 17, 21, 25, 29]$ \\
    \bottomrule
  \end{tabular}
\end{table}

\begin{table}[h]
  \caption{Hyperparameters: GNOCCHI}
 \label{tab:GNOCCHI_hps}
  \centering
  \begin{tabular}{llll}
    \toprule
    Parameter     &  Value \\
    \midrule
    Latent Dimensionality & $\{5, 15\}$ \\
    $\alpha_{\text{MMD}}$ & $-0.5$ \\
    $\lambda_{\text{MMD}}$ & $100$\\
    score loss weight & $0.7$ \\
    recon loss weight & $0.3$\\
    \# of diffusion steps & $\{200, 500\}$ \\
    $\beta$ range & $[0.001, 0.01]$ \\
    \midrule
    Auxiliary Encoder Hidden Size & $64$  \\
    Auxiliary Encoder Dropout  & $0.09$ \\
    Auxiliary Encoder Learning Rate & $0.0005$  \\
    Auxiliary Encoder Weight Decay & $0.000001$ \\
    Noise Predictor Hidden Size & $4096$  \\
    Noise Predictor Dropout  & $0.03$ \\
    Noise Predictor Learning Rate & $0.0005$ \\
    Noise Predictor Weight Decay & $0.0$ \\
    Noise Predictor Positional Embedding Size & $5$ \\
    \bottomrule
  \end{tabular}
\end{table}

\newpage
\section{LFADS \label{appendix: lfads}}
\subsection{Model architecture and training details}
For the artificial neural response experiments, LFADS was trained to minimize the mean-squared error between the model output and the RNN hidden states, with KL regularization against a multivariate normal prior with scaling given in Table \ref{tab:comparison-lfads-hps}. For the biological neural data experiment, LFADS was trained to minimize the Poisson negative log-likelihood of the spiking activity, given the model output, regularized by KL with scaling also given in Table \ref{tab:comparison-lfads-hps}. For each experiment, LFADS was applied using the same set of hyperparameters, with the initial condition dimensionality for the artificial random target task and biological random target task given in this order inside brackets in Table \ref{tab:comparison-lfads-hps}.

\subsection{Smoothing the input to GNOCCHI \label{lfads-smoothing}}
\begin{figure*}[h]
    \centering
    \includegraphics[width = \textwidth]{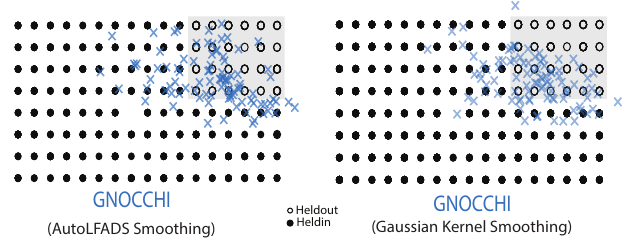}
    \vspace{-6mm}
    \caption{\textbf{Smoothing input to GNOCCHI with AutoLFADS vs. Gaussian kernel}. We show empirically that the prediction of held-out targets from GNOCCHI codes is not affected by whether the data are smoothed using AutoLFADS or a Gaussian kernel.}
    \label{fig:Supp--SmoothingControl}
\end{figure*}

For the biological neural data experiments, we used the smoothed spiking activity from an LFADS model as the input to the GNOCCHI-based model. This was due to the GNOCCHI-based model only currently having been validated for MSE-based reconstructions. The hyperparameters of the LFADS model are given in Table \ref{tab:lfads-for-smoothing-hps}. These hyperparameters were selected according to Binary tournament population-based training \cite{lfads-torch}, as in the AutoLFADS framework \cite{AutoLFADS}. As a control on whether the held-out generalization results shown in Figure \ref{fig:NeuralData}E are dependent on the method of smoothing for the input to GNOCCHI, we compare to a GNOCCHI model trained on spikes smoothed with a Gaussian kernel (s.d. 4 bins or 80ms). The results shown in Figure \ref{fig:Supp--SmoothingControl} demonstrate that held-out prediction of target location is comparable between smoothing approaches.

\begin{table}[h]
  \caption{Hyperparameters: LFADS for smoothing GNOCCHI input}
 \label{tab:lfads-for-smoothing-hps}
  \centering
  \begin{tabular}{lll}
    \toprule
    Parameter     & Value \\
    \midrule
    Initial Condition Encoder Dimensionality & 100 \\
    Initial Condition Dimensionality & 100 \\
    Controller Input Encoder Dimensionality & 80 \\
    Controller Input Lag & 1 \\
    Controller Dimensionality & 80 \\
    Controller Output Dimensionality & 4 \\
    Generator Dimensionality & 100 \\
    Factor Dimensionality & 40 \\
    Coordinated Dropout \cite{CoordinatedDropout} Rate & $0.39$ \\
    Initial Condition Prior & Multivariate Normal \\
    Initial Condition Prior Mean & 0 \\
    Initial Condition Prior Variance & 0.1 \\
    Controller Output Prior & Auto-Regressive Multivariate Normal \\
    Controller Output Prior $\tau$ & 10 \\
    Controller Output Prior Variance & 0.1 \\
    Dropout & $0.061$ \\
    RNN cell clip & 5.0\\
    Learning Rate & $2.1 \times 10^{-3}$ \\
    Adam $\beta_1$ & 0.9 \\
    Adam $\beta_2$ & 0.999 \\
    Weight Decay & 0.0 \\
    KL on initial conditions & $2.8 \times 10^{-6}$ \\
    KL on controller output & $5.6\times 10^{-7}$ \\
    $L_2$ penalty weight on generator & $0.45$ \\
    $L_2$ penalty weight on controller & $0.076$ \\
    \bottomrule
  \end{tabular}
\end{table}

\subsection{Hyperparameter selection}
The default parameters specified in \cite{lfads-torch} were used initially, and we found that these parameters work well across each dataset. As a result, we have continued with these parameters, changing only the initial condition dimensionality and turning off use of the controller network to ensure comparable usage to GNOCCHI when doing latent navigation. To select the KL penalty, we did a sweep across 4 values spanning the range $[1\times 10^{-8}, 1\times 10^{-4}$] across 5 random seeds, choosing the top performing model according to validation set reconstruction performance. The parameters used in each experiment are summarized in Table \ref{tab:comparison-lfads-hps}.

\begin{table}[h]
  \caption{Hyperparameters: LFADS}
 \label{tab:comparison-lfads-hps}
  \centering
  \begin{tabular}{lll}
    \toprule
    Parameter     & Value \\
    \midrule
   Initial Condition Encoder Dimensionality & 100 \\
    Initial Condition Dimensionality & $\{5,15\}$ \\
    Generator Dimensionality & 100 \\
    Factor Dimensionality & 40 \\
    Coordinated Dropout \cite{CoordinatedDropout} Rate & 0.3 \\
    Initial Condition Prior & Multivariate Normal \\
    Initial Condition Prior Mean & 0 \\
    Initial Condition Prior Variance & 0.1 \\
    Dropout & 0.02 \\
    RNN cell clip & 5.0\\
    Learning Rate & $1.0\times 10^{-2}$ \\
    Adam $\beta_1$ & 0.9 \\
    Adam $\beta_2$ & 0.999 \\
    Weight Decay & 0.0 \\
    KL on initial conditions & $\{1\times 10^{-8}, 1 \times 10^{-5}\}$ \\
    $L_2$ penalty weight on generator & 0.0 \\
    $L_2$ penalty weight on controller & 0.0 \\
    \bottomrule
  \end{tabular}
\end{table}

\newpage
.
\newpage

\section{Compute resources}
We used an internal computing cluster consisting of a combination of NVIDIA GeForce 2080 RTX Ti and A40 GPUs. The single model runtime for GNOCCHI and LFADS is approximately 30 minutes wall time, and the model parameter file sizes are roughly 450 MB and 0.66 MB, respectively.

\section{Open source packages used}
\begin{itemize}
    \item \texttt{torch} \cite{PyTorch} (BSD license)
    \item \texttt{pytorch-lightning} \cite{PyTorch_Lightning} (Apache 2.0 license)
    \item \texttt{ray.tune} \cite{ray_tune} (Apache 2.0 license)
    \item \texttt{scikit-learn} \cite{scikit-learn} (BSD license) 
    \item \texttt{lfads-torch} \cite{lfads-torch} (Apache 2.0 license) 
    \item \texttt{PyTorch-VAE} \cite{PyTorch-VAE} (Apache 2.0 license)
    \item \texttt{hydra} \cite{Yadan2019Hydra} (MIT License)
\end{itemize}

\end{document}